\documentclass{article}

\usepackage{microtype}
\usepackage{graphicx}
\usepackage{subfigure}
\usepackage{booktabs} 

\usepackage{enumitem}
\usepackage{amssymb}
\usepackage{amsmath}

\usepackage{threeparttable}
\usepackage[dvipsnames]{xcolor}
\usepackage{tabularx}
\usepackage{comment}
\newcommand\numberthis{\addtocounter{equation}{1}\tag{\theequation}}

\usepackage{hyperref}

\usepackage[accepted]{icml2021}

\icmltitlerunning{Optimal Relevance Mapping for Continual Learning}

\begin{document}

\twocolumn[
\icmltitle{Understanding Catastrophic Forgetting and Remembering in \\Continual Learning with Optimal Relevance Mapping}



\icmlsetsymbol{equal}{*}

\begin{icmlauthorlist}
\icmlauthor{Prakhar Kaushik}{to}
\icmlauthor{Alex Gain}{to}
\icmlauthor{Adam Kortylewski}{to}
\icmlauthor{Alan Yuille}{to}
\end{icmlauthorlist}

\icmlaffiliation{to}{Department of Computer Science, Johns Hopkins University, USA}

\icmlcorrespondingauthor{Prakhar Kaushik}{pkaushi1@jh.edu}

\icmlkeywords{Machine Learning, ICML}

\vskip 0.3in
]



\printAffiliationsAndNotice{}  

\begin{abstract}
Catastrophic forgetting in neural networks 
is a significant problem for continual learning. 
A majority of the current methods 
replay previous data during training, which
violates the constraints of an ideal continual learning system. 
Additionally, current approaches that deal with forgetting ignore the problem of catastrophic remembering, i.e. the worsening ability to discriminate between data from different tasks.
In our work, we introduce Relevance Mapping Networks (RMNs) which are inspired by the Optimal Overlap Hypothesis. 
The mappings reflects the relevance of the weights for the task at hand by assigning large weights to essential parameters.
We show that RMNs learn an optimized representational overlap that overcomes the twin problem of catastrophic forgetting and remembering.
Our approach achieves state-of-the-art performance across all common continual learning datasets, even significantly outperforming data replay methods while not violating the constraints for an ideal continual learning system. 
Moreover, RMNs retain the ability to detect data from new tasks in an unsupervised manner, thus proving their resilience against catastrophic remembering.

\end{abstract}
\section{Introduction}\label{intro}
Continual learning refers to a learning paradigm where different data and tasks are presented to the model in a sequential manner, akin to what humans usually encounter. But, unlike humans or animal learning, which is largely incremental and sequential in nature, artificial neural networks (\textit{ANNs}) prefer learning in a more concurrent way and have been shown to forget \textit{catastrophically}. The term \textit{catastrophic forgetting (CF)} in neural networks is usually used to define the inability of \textit{ANNs} to retain old information in the presence of new one.

\textbf{Continual Learning (CL) in Neural Networks.} \label{cl}
The widely understood formulation of continual learning
 refers to a learning paradigm where \textit{ANNs} are trained \textit{strictly} sequentially on different data and tasks~\cite{chen2018lifelong, mundt2020wholistic}. The important conditions of the training paradigm are:
\begin{enumerate}[topsep=0pt,itemsep=-1ex,partopsep=1ex,parsep=1ex]
    \item Sequential Training i.e. for a \textit{single} neural network $f$ with parameters $\theta$ trained at time $\mathcal{T}$ with sequentially available data $\mathcal{D}_{1...N}$,\\
        $\mathcal{T}_1[f_{\theta}(\mathcal{D}_1)] < \mathcal{T}_2[f_{\theta^*}(\mathcal{D}_2)] < ... < \mathcal{T}_N[f_{\theta^{**}}(\mathcal{D}_\mathcal{N})]$
    \item No negative exemplars, examples or feedback i.e. future (or past) data samples cannot be provided to the network with the current data/task.\\
    $(\mathcal{D}_1\cap\mathcal{D}_T)\cup...\cup(\mathcal{D}_{T-1}\cap\mathcal{D}_T)\cup(\mathcal{D}_T\cap\mathcal{D}_{T+1})\cup(\mathcal{D}_T\cap\mathcal{D}_{T+2})...\cup(\mathcal{D}_T\cap\mathcal{D}_{T+N}) = \emptyset$
\end{enumerate}
Despite this formulation, since \cite{robins_catastrophic_1993} showed the promise of memory replay methods in dealing with \textit{CF} and the prevalence of cognitive/neuro science inspired theories regarding memory, it is of no surprise that \textit{rehearsal/replay buffer/generative replay} methods dominate the current state of the art (SOTA) benchmarks~\cite{titsias2019functional, fear17, pan2021continual, lee_continual_2020}. However, they clearly \textit{violate} the conditions of the \textit{CL} paradigm. 
Additionally, some most prominent current \textit{CL} methods~\cite{Serr2018OvercomingCF, lee_continual_2020, guo2020improved, yoo_snow_2020} change the \textit{ANN} altogether by adding new convolutional/linear layers for each task (for e.g. using multi heads -different last linear layer for each task- has become a common practice) or using a mixture of \textit{ANNs} which violates the above conditions as well since we are not training the same \textit{ANN} $f_{\theta}$ anymore. \textbf{\textit{In contrast, we aim to develop a learning paradigm that strictly obeys the formulation of CL without applying data replay or introducing new sets of additional ANN models or convolutional/linear layers during training and inference.}}

\textbf{Catastrophic Forgetting} is a direct implication of continual learning in \textit{ANNs} and is largely considered a direct consequence of the overlap of distributed representations in the network.
Most prior works deal with \textit{CF} by either completely removing the representational overlap \cite{French1991, kirkpatrick2017overcoming} or more frequently, 
by replaying data from previous tasks.
Data replay methods can deal with \textit{CF} but, in turn, lead to a reduced capability of the network to 
discriminate between old and new inputs \cite{doi:10.1080/09540099550039264}. This is referred to as \textbf{Catastrophic Remembering (CR)} (Refer to Section~\ref{cr} for a detailed discussion) and 
has been shown to be a significant limitation of replay methods \cite{robins_catastrophic_1993,doi:10.1080/09540099550039264}.

\textbf{The goal of this work} is that we attempt to develop a method for continual learning for deep neural networks which can alleviate the twin problem of Catastrophic Forgetting and Catastrophic Remembering at the same time, without violating or relaxing the conditions of a \textit{strict} continual learning framework.

Our proposed approach builds on the following \textbf{Optimal Overlap Hypothesis}:
 \textit{For a strictly continually trained deep neural network, 
 catastrophic forgetting and remembering can be minimized, without additional memory or data, by learning optimal representational overlap, such that the representational overlap is reduced for unrelated tasks and increased for tasks that are similar.}\par
More formally, for an \textit{ANN} $f_{\Theta}(D)$ with parameter $\Theta$ and sequentially available data $D_i$ over number of tasks/data $i \in [ 1,\mathbb{T}]$ instead of trying to enforce \textit{over-generalization} which is to learn a superset parameter space which encompasses the sequential tasks \{$\theta_i\mid \cup\theta_{1...\mathbb{T}}=\Theta \wedge \cap \theta_{1... \mathbb{T}} = \emptyset$\} or complete separation of weight space \{$\theta_i\mid \cup\theta_i \subsetneq \Theta$\} 
, we try and learn optimal overlaps amongst the sequentially learned parameter sets. That means \{$\forall{i,j}\in \mathbb{T}\ni i \neq j \mid \theta_i \cap \theta_j = \mathcal{A} \wedge \cup\theta_i = \Theta$\} where $\mathcal{A}\in [\emptyset,\theta_{i/j}]$.\\
Inspired by this hypothesis, we propose \textit{Relevance Mapping} for continual learning to alleviate \textit{CF} and \textit{CR}. 
During the continual learning process, our method learns the neural network parameters and a task-based relevance mask on the hidden layer representation concurrently. The almost-binary relevance mask keeps a portion of the neural network weights static and hence is able to maintain the knowledge acquired from previous tasks, while the rest of the network adapts to the new task.
Our experiments demonstrate that Relevance Mapping Networks outperform all related works by a wide margin on many popular continual learning benchmarks (Permuted MNIST, Split MNIST, Split Omniglot, Split CIFAR-100), hence alleviating catastrophic forgetting without relaxing or violating the conditions of a \textit{strict} continual learning framework.
Moreover, we demonstrate that Relevance Mapping Networks are able to detect new sequential tasks in an unsupervised manner with high accuracy, hence alleviating catastrophic remembering.
\\

\textbf{In summary, our contributions are:}
\begin{itemize}[topsep=0pt,itemsep=.5ex,partopsep=1ex,parsep=1ex]
    \item We introduce Relevance Mapping Networks which learns binary relevance mappings on the weights of the neural network concurrently to every task. We demonstrate that our model efficiently deal with the twin problem of catastrophic forgetting and remembering.
    \item \textbf{Our method achieves \textit{SOTA} results on all popular continual learning benchmarks} without relaxing the conditions of a strict continual learning framework.
    \item We re-introduce the concept of \textit{Catastrophic Remembering} for deep neural networks and show that our method is capable of dealing with the same (becoming the first modern methodology to elevate catastrophic forgetting and remembering concurrently). 
\end{itemize}

\section{Related Work}\label{sec:related}

\textbf{Continual Learning.} Current continual learning mechanisms dealing with \textit{CF} are broadly classified into \textit{regularization approaches, dynamic architecture, complementary learning systems and replay architectures} \cite{parisi2018continual}. Primarily based on the \textit{Stability-Plasticity Dilemma} \cite{merm13} concept, regularization approaches impose constraints on weight updates to alleviate catastrophic forgetting like \textit{Elastic Weight Consolidation} (EWC) \cite{kirkpatrick2017overcoming} and \textit{Learning Without Forgetting} \cite{Li_2018}. These methods do not ordinarily violate the conditions of the \textit{CL} framework but have been shown to suffer from brittleness due to representational drift \cite{titsias2019functional, kemker2017measuring} and thus are usually combined with other methods. Rehearsal/replay buffer methods, like \cite{titsias2019functional} which are the state-of-the-art methods, use a memory store of past observations to remember previous tasks in order to alleviate the brittleness problem. However, these are not representative of \textit{strict sequential learning} insofar that they still require re-learning of old data to some extent and perform significantly worse the less samples are replayed, and they may struggle to represent uncertainty about unknown functions.

There are no known methods which deal with \textit{CR} in continual learning framework, with our method being the \textit{first of its kind} to be able to combat \textit{catastrophic forgetting and remembering} in a strict continual learning framework.

\textbf{Catastrophic Remembering} refers to the tendency of artificial neural networks to abruptly lose the ability to discriminate between old and new data/task during sequential learning. It is an important problem and inherently attached to the problem of catastrophic  forgetting. But, unlike the problem of catastrophic forgetting, which has a rich literature of research, catastrophic remembering has not been explored 
outside of minor discussions in early works \cite{shark95, LEWANDOWSKY1995329, French1991}. In this work, we discuss CR from a probabilistic perspective (Section \ref{cr}) and demonstrate that related work suffers from CR in our experiments in Section \ref{cr-ex}. Finally, we demonstrate that our proposed Relevance Masking Networks are much more resilient to catastrophic remembering.

\textbf{Similar Methods.}\label{sim_meth}
The idea of using soft-masking in networks (usually on non-linear activations) has been utilized before in novel ways for solving different problems. However, few of them, if any, ground these methods in some underlying concept (Optimal Overlap in our case) and often these methods include masks which are mutually exclusive, for example, for sparsity learning~\cite{zhu2017prune}, joint learning~\cite{mallya} where \textit{piggyback} a pretrained network by using a non-differentiable mask thresholding function and value, etc. In contrast, we don’t require our models to be pretrained or be thresholded. In \textit{CL}, the following methods appear to be closest to our Relevance Mapping Networks (RMNs): 

\cite{Serr2018OvercomingCF} proposes hard attention (\textit{HAT}), a task based attention mechanism which can be considered the most similar to our \textit{RMN}. It differs from \textit{RMN} due to following reasons- 
(i) They utilize task embeddings and a positive scaling parameter - and a gated product of these two is used to produce a non-binary mask - unlike our \textit{RMNs} which don't use either a task embedding or a scaling parameter and is necessarily binary.
(ii) Unlike \textit{RMNs}, the attention on the last layer in \textit{HAT} is manually hard-coded for every task.
(iii) A recursive cumulative attention mechanism is employed to deal with multiple non binary mask values over tasks in \textit{HAT}. \textit{RMNs} however have no need for such a mechanism.
(iv) \textit{HAT} cannot be used in a unsupervised \textit{CL} setup or to deal with \textit{CR} and has not been implemented with more complex network architectures like Residual Networks.

\cite{jung2020continual} uses \textit{proximal gradient descent algorithm} to progressively freeze nodes in an \textit{ANN}. 
(i) Unlike \textit{RMNs}, this method employs selective regularization to signify node importance (which is calculated by lasso regularization). 
(ii) This method progressively uses up the parameter set of the \textit{ANN} and it is unclear whether it can be used for an arbitrary large number of sequential tasks. 
(iii) This method is also unable to deal with unsupervised learning scenario or \textit{CR}. (iv) This method uses a different classification layer for each task - relaxing the core constraints of the problem altogether.

\cite{ferrari_memory_2018} (i) calculates the parameter importance by calculating sensitivity of the squared $l2$ norm of the function output to their changes and then uses regularization (similar to \cite{kirkpatrick2017overcoming} to enforce in sequential learning, unlike \textit{RMNs}. 
    (ii) The method enforces fixed synaptic importance between tasks irrespective to their similarity and unlike our work, doesn't seem to be capable of working under Unsupervised Learning scenarios.
    
\cite{yoo_snow_2020} propose \textit{SNOW} and (i) uses a unique channel pooling scheme to evaluate the channel relevance for each specific task which differs from \textit{RMN's} individual node relevance mapping strategy. 
    (ii) Importantly, this work, unlike \textit{RMNs}, employs a pre-trained model which is frozen source model which already \textit{overgeneralizes} to the \textit{CL} problem at hand and thus makes this method inapplicable for dealing with \textit{CR}. (iii) It also doesn't seem to be capable of handling unsupervised learning/testing scenarios.

\label{related}

\section{Strict Continual Learning and a Catastrophic Memory}\label{disc}
Continual learning has been an important topic since the onset of Machine Learning and the fact that \textit{ANNs} are incapable of learning continually due to Catastrophic Forgetting has been a significant drawback. 
\textit{CF} has been strongly identified with overlap of distributed representations \cite{French1991}.

\textbf{Catastrophic forgetting from a probabilistic view.}
Intuitively, given a learnt initial set of parameters $\theta_i$ for a neural network $f$ and a task $i$ with data $D_i$, the network's parameters get overwritten
when it learns a new set of network parameters $\theta_{i+1}$ from new data $D_{i+1}$ for the $(i+1)_{th}$ task. 
To facilitate the conceptual understanding of CF, we consider continual learning from a probabilistic perspective, where  optimizing the parameters $\Theta$ of $f$ is tantamount to finding their most probable values given some data $\mathcal{D}\mid\mathcal{D}\supset D_1,...,D_n$  \cite{kirkpatrick2017overcoming}. 
We can compute the conditional probability of the first task $\mathcal{P}(\theta_1|D_1)$ from the prior probability of the parameters $\mathcal{P}(\theta_1)$ and the probability of the data $\mathcal{P}(D_1|\theta_1)$ by using Bayes’ rule. Hence, for the first task,
\begin{eqnarray}
\begin{aligned}
    \log \mathcal{P}(\theta_1|D_1) {=} \log \mathcal{P}(D_1|\theta_1){+}\log\mathcal{P}(\theta_1)
    {-}\log\mathcal{P}(D_1).\label{eq:in}
\end{aligned}
\end{eqnarray}
Note that the likelihood term $\log \mathcal{P}(D_1|\theta_1)$ simply represents the negative of the loss function for the problem at hand \cite{kirkpatrick2017overcoming}. Additionally, the posterior term is usually intractable and only approximated for \textit{ANNs} \cite{titsias2019functional, nguyen2017variational, kirkpatrick2017overcoming} and we are just considering it here without change for analysis purposes only.\\
If we were to now train the same network for a second task, the posterior from \eqref{eq:in} now becomes a prior for the new posterior. If no regularization or method is included to preserve the prior information, we'd optimize for the second task,
\begin{align*}
    \log \mathcal{P}(\theta_{1:2}|D_{1:2}) {=} &\log \mathcal{P}(D_2|\theta_2){+}\log\mathcal{P}(\theta_1|D_1)\numberthis\label{eq:2}\\
    &{-}\log\mathcal{P}(D_2)\\
    {=} &\log\mathcal{P}(D_2|\theta_2){+}\log\mathcal{P}(D_1|\theta_1)\numberthis\label{eq:3}\\
    &{+}\log\mathcal{P}(\theta_1){-}\log\mathcal{P}(D_1){-}\log\mathcal{P}(D_2).
\end{align*}
If the likelihood term is not optimised over both $\theta_1$ and $\theta_2$, as would happen normally in a ordinary \textit{ANN} training setup, 
the prior information can be overwritten, leading to the condition commonly  referred to as catastrophic forgetting.

\textbf{Overcoming catastrophic forgetting.} 
We can clearly see from Eq. \ref{eq:3}, if we had access to the previous data $D_1$ or if both $\theta_1$ and  $\theta_2$ were independent of one another, we could approximate a well optimized posterior. However, in a typical continual learning setting we do not have access to the previous data and cannot ordinarily make independence assumptions on the sequentially learnt parameters. However, this provides us a with a crucial conceptual understanding of how to deal with \textit{CF} and an insight towards understanding the mechanisms of current popular \textit{CL} methods, which aim to overcome either of the two mentioned restrictions.
In particular, some recent works \cite{mallya, Serr2018OvercomingCF, jung2020continual} try to effectively separate the model parameters $\theta_i$ for different tasks, as initially proposed by \cite{French94dynamicallyconstraining}. 
The most successful recent works~\cite{pan2021continual, titsias2019functional, Chaudhry_2018_ECCV, guo2020improved, fear17} involve data replay methods, which relax the previous data availability restriction and have been shown to be effective initially by ~\cite{robins_catastrophic_1993}.
Despite its success in dealing with CF to an extent, data replay drastically diminishes the discriminative ability of the \textit{ANN}, which is referred to as \textit{Catastrophic Remembering} \cite{robins_catastrophic_1993}. This usually happens when the \textit{ANN} learns a more general function $f(\Theta)$ than necessary,  generalizing 
not only to the individual tasks, but to the entire sequential set of tasks \{$\theta\mid \forall\theta_i \subsetneq \Theta$\} which we has been referred to as \textit{overgeneralization} \cite{robins_catastrophic_1993}. 
The network can then experience a sense of \textit{extreme deja vu} \cite{doi:10.1080/09540099550039264} and is unable to differ the old from new data. 
\subsection{Catastrophic Remembering}\label{cr}


For a better understanding of \textit{CR} and why \textit{CF} alleviation aggravates it, we calculate the posterior after the $n_{th}$ task learnt continually using Eq. \eqref{eq:3},
\begin{eqnarray}
    \begin{aligned}
    \log \mathcal{P}(\theta_{1:n}|D_{1:n}) =   &\log \mathcal{P}(D_n|\theta_n)+\sum_{i=1}^{n-1}\log \mathcal{P}(D_i|\theta_i)\\
                                                 &+\log\mathcal{P}(\theta_1)-\mathcal{C} \label{eq:6}\\
    \end{aligned}
\end{eqnarray}
where $\mathcal{C}$ is a constant representing the sum of the normalization constants $\sum_{i=1}^n\log \mathcal{P}(D_i)$. As discussed earlier, the information of from the previous tasks is passed to the next sequential task as a prior ($\sum_{i=1}^{n-1}\log \mathcal{P}(D_i|\theta_i)$). The problem of loss of discriminative ability arises when for an arbitrary large $n$ when the prior term far exceeds the currently optimized likelihood. For Eq. \eqref{eq:6}, that means
\begin{eqnarray}
    \begin{aligned}
    \log \mathcal{P}(D_n|\theta_n) << \sum_{i=1}^{n-1}\log \mathcal{P}(D_i|\theta_i).
    \end{aligned}
\end{eqnarray}
In the context of data replay methods this intuitively means that if the number of data from the previous tasks $\{D_1,\dots,D_{n-1}\}$ is far bigger than the data in the current task $D_n$
the contribution of the present likelihood to the posterior is negligible and no new features are learnt by the \textit{ANN} to account for the new dataset/task. 
This, in turn, gives the model a sense of false \textit{familiarity} with a new input and the model is no longer able to discriminate between old and new inputs. 
The above explanation, though not exhaustive, provides a initial understanding from a Bayesian viewpoint. 

One may argue against the necessity of the discriminative property that \textit{CR} attacks in \textit{ANNs}. While, it is true that just concentrating on generalization may allow us to ignore the problems of \textit{CR}, but novel input and task detection
are important problems in Artificial Intelligence, Computer Vision and Robotics.
It can be necessary to detect new inputs to learn more robust features for the current data, e.g. a self driving network may need to identify whether it is familiar with a current set of input data. 
Additionally, \textit{Recognition and discrimination memory} are important aspects of human memory and learning - concepts which artificial networks have been trying to replicate. 

\textbf{Balancing Forgetting and Remembering}
Having gained a basic understanding about \textit{CF} and \textit{CR}, an astute reader realizes the crux of the problem that we are dealing with. 
Alleviating \textit{CF} appears to aggravate \textit{CR}.
While current literature focuses on alleviating CF, the problem of CR does not receive much attention. 
One aim of this work is to shed light on the twin problem of catastrophic forgetting and remembering and to introduce a 
method which can balance alleviating both problems concurrently. 

\section{Relevance Mapping for Continual Learning}\label{amm}
We introduce \textit{Relevance Mapping}, which is a method inspired by the \textit{Optimal Overlap Hypothesis}, that aims to learn an optimal representational overlap, such that unrelated tasks use different network parameters, while allowing similar tasks to have a representational overlap. 
Note that our method avoids data replay and instead aims to achieve independence between network weights that are used for different sequential tasks.

\textbf{Algorithmic implementation of Relevance Mapping.}
To illustrate and motivate Relevance Mapping Networks (RMNs) using a simple example, we consider a multilayer perceptron (MLP) with two layers $f$ 
defined as 
\begin{eqnarray}
	f(x) \triangleq \sigma(W_2 \sigma( W_1 x)),
\end{eqnarray}
where $x \in \mathbb{R}^{d_1}$, $W_1 \in \mathbb{R}^{d_2 \times d_1}$, and $W_2 \in \mathbb{R}^{d_3 \times d_2}$, and $\sigma$ denotes a nonlinear activation function. We denote the set of weights as $\mathbf{W} \triangleq \{ W_1, W_2 \}$. Although it may depend on the dimensionality of the task, overparameterization occurs even in these simple MLP settings. For a sufficiently simple task, only a subset of the parameters in $\mathbf{W}$ are often required~\cite{frankle2019lottery}. For example, if the optimization task has ground truth outputs specified as $f^*(x) =  \sigma(W_2^* \sigma( W_1^* x))$ for optimized weights $\{ W_1^*, W_2^* \} $, and $||W_1^*||_0 + ||W_2^*||_0 \ll d_3 d_2 + d_2 d_1$ (i.e. the number of non-zero weights needed for the ground-truth function is much less than the number of total weight parameters) then only $||W_1^*||_0 + ||W_2^*||_0$ weight parameters are necessary to be learned in network $f$. In theory, if we could learn the \textit{importance} or \textit{relevance} of each weight node, we could apply a zero-mask to the non-essential parameters without pruning or modifying them and still successfully learn the ground-truth. A set of mappings can be denoted as $\mathbb{M_P} = \{ \mathbb{M_P}_1, \mathbb{M_P}_2 \}$, where $\mathbb{M_P}_1 \in \{0, 1\}^{d_2 \times d_1}$ and $\mathbb{M_P}_2 \in \{0, 1\}^{d_3 \times d_2}$, explicitly representing the neuron-to-neuron connections of the network.
The initialized relevance mappings of an \textit{ANN} can be approximated by a logit-normal distribution mixture which is rounded during inference. 
\begin{equation*}
    \mathbb{M_P}_k \approx \prod_k \mathcal{L}_R\mathcal{N}(\mu_k,\,\sigma_k^{2})
\end{equation*} where $\mu,\sigma$ are the initializing distribution parameters and $\mathcal{L}_R$ sigmoidal pseudo-round function: \begin{equation}
\mathcal{L}_R(x_k; \beta) = \frac{1}{1 + \exp(-(\beta (x_k - 0.5)))}
\end{equation}
This is done in order to make the mappings differentiable and the individual mixture components are jointly optimized for the task with the network parameters.


In theory, \textit{any} network $f$ with weight tensors $\mathbf{W}$ can have such corresponding sets of neuron connection representations $\mathbb{M_P}_1,\mathbb{M_P}_2, \dots \mathbb{M_P}_T$ for $T$ tasks/mappings, where each set $\mathbb{M_P}_i$ activates a subnetwork mapping in $f$ that could be used for various purposes for a task $i$.


Note that $\lim_{\beta \rightarrow \infty} (\mathcal{L}_R(x, \beta))$ for $x \in [0,1]$ is equivalent to the rounding function. Here, $\beta$ is a learnable, layer-wise parameter (i.e., in our implementation, there is one specific $\beta$ for every layer of a given network) that controls the ``tightness'' of $R$. To achieve an approximate neuron-connection representation, we define $\hat{\mathbb{M_P}} = \mathcal{L}_R(\mathbb{M_P}; \beta)$ where $\mathbb{M_P}$ is initialized from some distribution with support $[0,1]$ (in experiments, we initialize $\hat{\mathbb{M_P}}$ with a clipped, skewed normal distribution). 

In our presented work, we can think of the \textit{RMNs} as replacing the weights of a network with the product of the weights and a binary relevance mixture. 
In this work, we introduce two algorithms, \textbf{Algorithm 1 and 2 (Supplementary Sec. 2)} which make use of Relevance Mapping. The former is used for traditional \textit{Supervised CL experiments} which used to evaluate \textit{CF} alleviation. The latter is used for the \textit{Unsupervised scenario} (new task detection and unsupervised task inference) concerning evaluation of \textit{CR} alleviation. Importantly, neither of the algorithms relax the conditions of a \textit{strict} CL framework (Section~\ref{cl}).\footnote{Refer to Supplementary for further method details}

\begin{table*}[t]
  \caption{Results on sequential learning tasks for the Split-MNIST \textit{(S-MNIST)}, Permuted-MNIST \textit{(P-MNIST)}, Sequential Omniglot \textit{(S-Omniglot)}, Split Cifar-100(20 tasks) with Resnet18 \textit{(RES-CIFAR)} and Split Cifar-100(5 tasks) \textit{(S-CIFAR100)} tasks. 
  Mean test accuracy results with standard deviation over five trials are shown where applicable.}
  \label{sequential-baselines}
  \vskip 0.15in
  \begin{center}
  \begin{small}
  \begin{sc}
  \begin{threeparttable}
  \begin{tabular}{lccccr}
    \toprule
    Algorithm & P-MNIST & S-MNIST & S-Omniglot & RES-CIFAR & S-CIFAR100\\
    \midrule
    VCL(\cite{nguyen2017variational})\tnote{$\mathcal{R}$,$\mathcal{H}$} & $90$ & $97$ & $53.86\pm2.3$& $-$ & $-$ \\
    & \textit{(200 pts/task)} & \textit{(40 pts/task)} & \textit{(3 pts/character)}\\
    HAT(\cite{Serr2018OvercomingCF})$_{\tnote{$\dagger$}}$ \tnote{$\mathcal{H}$} & $91.6$ & $99$ & $5.5 \pm 11.1$ & $23.6\pm8.8$ & $59.2 \pm0.7$\\
    RWALK(\cite{Chaudhry_2018_ECCV})\tnote{$\mathcal{R}$,$\mathcal{H}$} & $-$ & $82.5$ & $71.0\pm5.6$ & $70.1$ & $58.1 \pm 1.7$\\
    &  & & & \textit{(5000 samples)}  \\
    AGS-CL(\cite{jung2020continual})$_{\tnote{$\dagger$}}$\tnote{$\mathcal{H}$} & $-$ & $-$ & $82.8\pm1.8$ & $27.6\pm3.6$ & $64.1 \pm 1.7$\\
    FRCL(\cite{titsias2019functional})\tnote{$\mathcal{R}$} & $94.3 \pm 0.2$ & $97.8 \pm 0.7$ & $81.47 \pm 1.6$ & $-$ & $-$ \\
     & \textit{(200 pts/task)} & \textit{(40 pts/task)} & \textit{(3 pts/character)}\\
    MEGA-II(\cite{guo2020improved})\tnote{$\mathcal{R}$,**} & $91.21$ & $-$ & $-$& $66.12\pm1.94$\tnote{$\mathcal{M}$} & $-$ \\
    & \textit{(256 pts/task)} & & & \textit{(1300 pts/task)}\\
    SNOW(\cite{yoo_snow_2020})$_{\tnote{$\dagger$}}$\tnote{a} & $-$ & $-$ & $82.8 \pm 1.8$ & $-$ & $-$ \\
    FROMP(\cite{pan2021continual})\tnote{$\mathcal{R}$} & $94.9 \pm 0.1$ & $99.0 \pm 0.1$ & $-$& $-$ & $-$ \\
    & \textit{(40 pts/task)} & \textit{(40 pts/task)} \\
    \midrule\midrule
    DLP(\cite{esk03}) & $82$ & $61.2$ & $-$& $-$ & $-$ \\
    EWC(\cite{kirkpatrick2017overcoming}) & $84$ & $63.1$ & $67.43 \pm 4.7$\tnote{$\mathcal{H}$} & $42.67 \pm 4.24$\tnote{$\mathcal{H}$} & $60.2 \pm 1.1$\tnote{$\mathcal{H}$} \\
    SI(\cite{zenke2017continual}) & $-$ & $57.6$ & $54.9\pm16.2$ & $45.49\pm0.2$\tnote{$\mathcal{H}$} & $60.3\pm1.3$\tnote{$\mathcal{H}$}\\
    MAS(\cite{ferrari_memory_2018})$_{\tnote{$\dagger$}}$ \tnote{$\mathcal{H}$} & $-$ & $-$ & $81.4 \pm 1.8$ & $42\pm1.9$ & $61.5 \pm 0.9$\\
    \textbf{RMN (Ours)} & $\textbf{97.727} \pm \textbf{0.07}$ & $\textbf{99.5} \pm \textbf{0.2}$ & $\textbf{85.33} \pm \textbf{1.7}$ & $\textbf{80.01} \pm \textbf{0.9}$ & $\textbf{70.02} \pm \textbf{2.5}$ \\
    \bottomrule
    \end{tabular}
    \begin{tablenotes}[para]\footnotesize
    \item[$\dagger$] similar methods(Section~\ref{sec:related})
    \item[$\mathcal{p}$] uses pretrained network
    \item[$\mathcal{R}$] uses data replay buffer
    \item[$\mathcal{H}$] Multiheaded layer implementation
    \item[**] Not trained over all tasks
    \item[a] Additional model is used
    \end{tablenotes}
    \end{threeparttable}
    \end{sc}
    \end{small}
    \end{center}
    \vskip -0.1in
\end{table*}

\textbf{Probabilistic interpretation of Relevance Mapping.}
French introduced the method of \textit{context-biasing} in \yrcite{French94dynamicallyconstraining} which produces internal representations which are both well distributed and well separated to deal with \textit{CF}. \textit{RMN} preserves a similar idea of distribution and separability without constraining for an explicit representation separation amongst posteriors learnt for the sequential tasks. The separation, in turn, is provided by the relevance mappings. 
\begin{eqnarray}
    \begin{aligned}
    &\mathcal{P}(\theta_1,\mathbb{M_P}_1|D_1) \propto \mathcal{P}(D_1|\theta_{\mathbb{M_P}_1})\mathcal{P}(\theta_{\mathbb{M_P}_1}) \label{eq:12}
    \end{aligned}
\end{eqnarray}
The $1st$ task of the \textit{CL} problem presented in Eq. \eqref{eq:12} is similar to Eq. \eqref{eq:in} with relevance mappings introduced under the conditions of the algorithm presented. $\theta_{\mathbb{M_P}_i}$ represents only a subset of $\theta$ for which $\mathbb{M_P}_i=1$. For learning the second task we optimize
\begin{eqnarray}
    \begin{aligned}
    &\mathcal{P}(\theta_{1:2}, \mathbb{M_P}_2|D_{1:2}) \propto \mathcal{P}(D_2|\theta_{\mathbb{M_P}_2})\mathcal{P}(\theta_1, \mathbb{M_P}_1|D_1). \label{eq:13}
    \end{aligned}
\end{eqnarray}
In Eq. \eqref{eq:13}, the second term on the right doesn't contribute anything to the optimization over the second task due to the presence of independent relevance mappings which effectively disengages $\theta_{\mathbb{M_P}_1}$ from further tampering and the next task receives a slightly constrained prior distribution that we can refer to as $\theta_2^{''}$. The $\theta_{\mathbb{M_P}_1}$ parameter set is however still available to the second task. Eq. \eqref{eq:13} now becomes 
\begin{eqnarray}
    \begin{aligned}
    &\mathcal{P}(\theta_{1:2}, \mathbb{M_P}_2|D_{1:2}) \propto \mathcal{P}(D_2|\theta_{\mathbb{M_P}_2})\mathcal{P}(\theta_2^{''}) \label{eq:14}
    \end{aligned}
\end{eqnarray}
which is effectively now a problem of just jointly optimizing an \textit{ANN's} parameters $(\Theta, \mathbb{M_P}_2)$ without any dependence on the previous task's posterior. We have effectively decomposed the sequential task parameters.
There are three scenarios that may occur w.r.t the optimised parameters i.e. ($k$ represents the individual elements) (i) $\mathbb{M_P}_2^k = \mathbb{M_P}_1^k \Rightarrow \theta_{\mathbb{M_P}_1}^k=\theta_{\mathbb{M_P}_2}^k$  (i) $\mathbb{M_P}_2^k = 1$ \& $\mathbb{M_P}_1^k=0$ $\Rightarrow \{\theta_{\mathbb{M_P}_1}^k\cap\theta_{\mathbb{M_P}_2}^k=\emptyset\}$(iii) $\mathbb{M_P}_2^k = 0$ \& $\mathbb{M_P}_1^k=1$. All of these scenarios can be handled by \textit{RMNs} thanks to the $\mathcal{O}_2$ hypothesis.

For $n$ tasks, \eqref{eq:14} becomes,
\begin{eqnarray}
    \begin{aligned}
    &\mathcal{P}(\Theta, \mathbb{M_P}|D_{1:n}) \propto \prod_{i=1}^n{\mathcal{P}(D_i|\theta_{\mathbb{M_P}_i})\mathcal{P}(\theta_i^{''})} \label{eq:15}
    \end{aligned}
\end{eqnarray}
Looking at \eqref{eq:15} which is a basic Bayesian expression for a normal \textit{ANN}, we can now understand that $\mathcal{O}_2$ \textit{hypothesis} inspired \textit{RMN} algorithm is capable of learning well separated and well distributed internal representations thanks to the \textit{posterior decomposition} induced by our method. This takes care of the problem of \textit{CF} and since the parameters of the model are jointly optimized over both the \textit{RMN} parameters $\Theta$ and the relevance mappings $\mathbb{M_P}$, the network cannot overgeneralize to a specific task given only $\Theta$ which, in turn takes care of \textit{CR}.

The focus in \textit{RMNs} is not to force a zero representational overlap or just generalize to all the sequential tasks altogether but rather to utilize the \textit{over-parameterization} property of \textit{ANNs}~\cite{frankle2019lottery} and learn an optimal representational overlap for all tasks in the weight space - corroborating the \textit{Optimal Overlap Hypothesis}. Therefore, there's no constraint on the maps $\mathbb{M_P}$ to minimize the overlap with each other or a global loss function which takes in account of the loss of individual tasks.
The map $\mathbb{M_P}$ for each task helps define a subset of the final weight mapping of the \textit{ANNs}. This subset may be disjoint or overlapping with other $\mathbb{M_P}$ defined weight subsets. Since, all the sequential tasks' parameter mappings are subsets of the final weight mapping (with $\mathbb{M_P}$ defining the set relationship), we are able to alleviate both CF (the final mappings generalizes well for all the tasks) and CR (the $\mathbb{M_P}$ preserve the relationship between the global and individual parametric mappings).
\section{Experiments}\label{ex}

\subsection{Supervised CL (Testing Catastrophic Forgetting)}\label{cf-ex}

We evaluate \textit{RMNs} on supervised sequential learning tasks, which enables us to measure their ability to alleviate \textit{CF}. In this setup, the network is given data for learning one task followed by another. The challenge lies in retaining the performance on previous tasks even as new tasks are learned, hence alleviating catastrophic forgetting. This experimental framework is commonly used in \textit{CL} literature.  

\textbf{Setup.}  
We use standard baseline architectures, including \textit{CNNs} \cite{lecun1995convolutional}, Siamese Networks \cite{koch2015siamese}, and Residual Networks\cite{he2015deep} and apply Relevance Mapping to them (denoted as CNN-RMN, Siamese-RMN. Resnet18-RMN, etc.). For task-wise classification, we let the classification output $f(x; W, \mathbf{\mathbb{M_P}}_1, \dots \mathbf{\mathbb{M_P}}_T) \triangleq argmax_{i \in \{ 1\dots T \}}(\{f(x; W, \mathbf{\mathbb{M_P}_i})\}$ so that no task-specific information is utilized at inference time. We also augment the loss function with L1-norm penalty on $\mathbf{\mathbb{M_P}}$ masks and sum of overlap of $\mathbf{\mathbb{M_P}}_1, \dots \mathbf{\mathbb{M_P}}_T)$ to reward sparsity and optimal separation of weight spaces, respectively. We adhere to the \textit{strict} \textit{CL}(Section~\ref{cl}) framework in \textit{RMN} experiments. 

\begin{figure}
\begin{center}
\centerline{\includegraphics[width=\columnwidth]{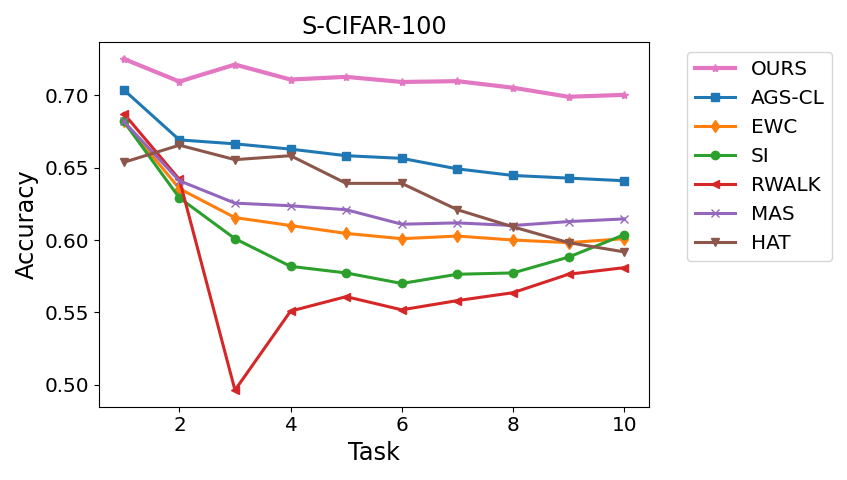}}
\vspace{-\baselineskip}
\caption{Average accuracy results on CIFAR-100 (10 tasks)}
\label{scifar}
\end{center}
\vspace*{-2.8\baselineskip}
\end{figure}

\textbf{Models.} As is common in related work \cite{titsias2019functional, kirkpatrick2017overcoming, nguyen2017variational, pan2021continual, jung2020continual}, we evaluate RMNs on five benchmarks: \textit{Permuted-MNIST} \cite{kirkpatrick2017overcoming}(P-MNIST), \textit{Split-MNIST}(S-MNIST), \textit{Sequential Omniglot}(S-OMNIGLOT) \cite{schwarz2018progress}, 10 task \textit{Split-Cifar100} \cite{zenke2017continual}(S-CIFAR100) and 20 task \textit{Split-Cifar100} (RES-CIFAR). To validate the efficacy of RMNs on complex architectures, the \textit{RES-CIFAR} model is trained on a Resnet18. For 10 task \textit{S-Cifar100}, we used 6 convolution layers followed by 2 fully connected layers (with ReLU activations). \textit{S-MNIST, P-MNIST} and \textit{S-Omniglot} architectures are same as in \cite{titsias2019functional}.

\textbf{Results and Discussion}
As seen in Table~\ref{sequential-baselines}, our \textbf{RMNs set the new state of the art across all continual learning benchmarks} presented, with improvements of $2.8\%$ (P-MNIST), $0.5\%$ (S-MNIST), $3.9\%$ (S-Omnliglot), $8.7\%$ (S-Cifar100) and $13.9\%$ (RES-CIFAR) over the previous SOTA. RMNs show their versatility in both simple (MLP) and complex (ResNet) architectures over both long (S-Omniglot, RES-CIFAR) and short continual learning, demonstrating their versatility. Figure~\ref{scifar} shows an example of the effectiveness of \textit{RMN}s as compared to other methods when dealing with \textit{CF}.

To keep comparisons fair amongst methods, Table~\ref{sequential-baselines} is divided into two parts by a 2-line separator. The above part includes all the methods which do not obey the conditions of a strict \textit{CL} framework.
Some compared methods\cite{nguyen2017variational, titsias2019functional, Chaudhry_2018_ECCV, guo2020improved, pan2021continual} employ data replay buffers while others cannot work efficiently without one or more multi-headed layers~\cite{nguyen2017variational, Chaudhry_2018_ECCV, jung2020continual} or a version of it - for e.g. \cite{Serr2018OvercomingCF} use manual hard coding of layers per task. The lower part of Table~\ref{sequential-baselines} consists of methods which are implemented 
in a \textit{strict} sequential learning setup.
Unlike most of the compared methods, \textbf{RMNs do not require any replay buffers, ensemble networks, meta networks, multi-headed layers or pretrained models} and yet are able to outperform methods which do use such methods.
We also compare RMNs with \textit{similar methods} as mentioned in Section~\ref{sim_meth} and see that RMNs substantially outperforms every one of them as seen in Table~\ref{sequential-baselines}.

\subsection{Unsupervised CL (Testing Catastrophic Remembering)}\label{cr-ex}

\textbf{Measuring Catastrophic Remembering}
For a good measure of CR, we need to evaluate how well a sequentially trained ANN discriminates between old and new data as well as how well does it discriminate between all the tasks/data after it has been trained on all of them.
To that end, we propose two tests: (1) (Unsupervised) New Task/Data Detection and (2) Unsupervised Task Inference.
In the new task detection setup, the \textit{ANN} is given no supervision with respect to the new data or task and has to detect this change. The model's performance (for e.g., the accuracy for each task in case of classification) is compared with the supervised continual learning version. In the second test, a trained model has to detect which specific task does the \textit{test input data} belongs to. A point to be noted is that the preconditions of the sequential training paradigm mentioned in Section~\ref{cl} are to be \textit{strictly} observed. The tests are mutually inclusive in terms of representing effectiveness w.r.t \textit{CR} alleviation i.e. a method should perform well on both tasks to be good candidate for fixing \textit{Catastrophic Remembering} (while still being able to alleviate \textit{Catastrophic Forgetting}).

\subsubsection{New Task/Data Detection}

\textbf{Setup.} The model is given no information about the tasks during training (and inference) time. The results are then compared with the full supervised version (Table \ref{sequential-baselines}). The performance degradation from the supervised learning results allow us to evaluate how well the model can alleviate CR. Here, we assume that no information of task labels is given, including the number of disparate tasks. 

\textbf{RMN Methodology.} In this case, we initialize $f$ with only a single $\mathbb{M_P}$, i.e. only a single forward inference path can be learned
at initialization, as seen in Line 2 of Algorithm 2 (Suppl. Sec. 2). We set the current task indicator as $est_j {=} 0$. Then, for each minibatch $x$ encountered, we run a task-switch-detection $(TSD)$ method, denoted as $TSD(x)$ which returns a boolean value. If $TSD(x)$ returns True, then $est_j$ is incremented and another set of $\mathbb{M_P}$ is added to $f$. 
We use a \textit{Relevance} modified Welsh’s t-test on the KL divergence between prior and posterior distributions of the model to determine a task switch \cite{titsias2019functional, hendrycks2016baseline, lee2018simple}.

\textbf{Results and Discussion.} 
Few methods~\cite{titsias2019functional, lee_continual_2020, pan2021continual} have tried to effectively deal with the harder problem of learning continually without task labels and none of these follow a \textit{strict} \textit{CL} framework. \cite{titsias2019functional} and \cite{pan2021continual} both employ a data replay buffer whereas \cite{lee_continual_2020} uses generative replay and a mixture of expert models (which leads to a large increase in computational and memory requirements).

\begin{table}[t]
\caption{Continual Learning without Task Labels.}
\label{unsup-clean}
\vskip -\baselineskip
\begin{center}
\begin{small}
\begin{sc}
\begin{threeparttable}
\vspace*{-\baselineskip}
\begin{tabular}{lccc}
\toprule
Algorithm & P-MNIST & S-MNIST & S-Omniglot\\
\midrule
FRCL\tnote{$\mathcal{R}$} & $94.3 \pm 0.2$ & $97.8 \pm 0.7$ & $81.47 \pm 1.6$ \\
FROMP\tnote{$\mathcal{R}$} & $94.9 \pm 0.1$ & $99.0 \pm 0.1$ & $-$\\
CN-DPM\tnote{$\mathbb{C},\mathcal{R}$} & $-$ & $97.53 \pm 0.3$ & $-$\\
\textbf{RMN (Ours)} & $97.73 \pm 0.1$ & $99.5 \pm 0.2$ & $85.33 \pm 1.7$ \\
\bottomrule
\end{tabular}
\begin{tablenotes}[para]\footnotesize
    \item[$\mathbb{C}$] \cite{lee_continual_2020}
    \item[$\mathcal{R}$] uses data replay buffer
\end{tablenotes}
\end{threeparttable}
\end{sc}
\end{small}
\end{center}
\vskip -\baselineskip
\end{table}

The usual methodology that is followed in an Unsupervised \textit{CL} learning setup involves \textit{boundary detection} between current and new tasks. Table~\ref{unsup-clean} shows the results of this setup amongst all the relevant methods. \textbf{RMNs achieve the state of the art for Unsupervised CL Learning} without the usage of data replay buffer, mixture of expert models or any kind of generative replay, unlike~\cite{titsias2019functional, pan2021continual, lee_continual_2020}.
Ordinarily, these task detection methods employ statistical tests like Welch's T-test over clean batches of data (the entire batch data belongs to either the current or the next task). This methodology fails when the incoming previous and the incoming batch are noisy with the incoming batch consisting of the new task data as well as old data. \textit{RMN}s however can easily deal with this by filtering the incoming batch via final layer activations - only if \textit{RMNs} have seen the data do they have high and confident activations before calculating the KL divergence test between the prior and posterior to detect the presence of a new set of data. However, none of the methods~\cite{titsias2019functional, lee_continual_2020} mentioned are capable of deploying over such a noisy data setup and are unable to learn continually. \cite{lee_continual_2020} does employ a \textit{Fuzzy Testing} scenario for Split-MNIST in which there are transition phases between tasks where the amount of new data increases linearly in each batch. Comparison on the same experiment is presented in Table~\ref{noisycrsplit}.

\begin{table}
\caption{Fuzzy Unsupervised Learning.}
\label{noisycrsplit}
\begin{center}
\begin{small}
\begin{sc}
\begin{tabular}{lcc}
\toprule
Algorithm &  S-MNIST \\
\midrule
CN-DPM\tnote{$\mathbb{C},\mathcal{R}$} & $93.22 \pm 0.07$ \\
\textbf{RMN (Ours)} & $99.1 \pm 0.5$ \\
\bottomrule
\end{tabular}
\end{sc}
\end{small}
\end{center}
\vskip -0.27in
\end{table}

\subsubsection{Unsupervised Task Inference} 

Under this \textit{novel} setup, the algorithm has to identify at inference time which task a data input belongs to amongst all the tasks it has learned. From a practical point of view, knowing which task in the sequential task list does the current inference data element belongs to, without human intervention opens up huge opportunities for automation and analysis. 
\begin{figure}[ht]
\begin{center}
\centerline{\includegraphics[width=\columnwidth]{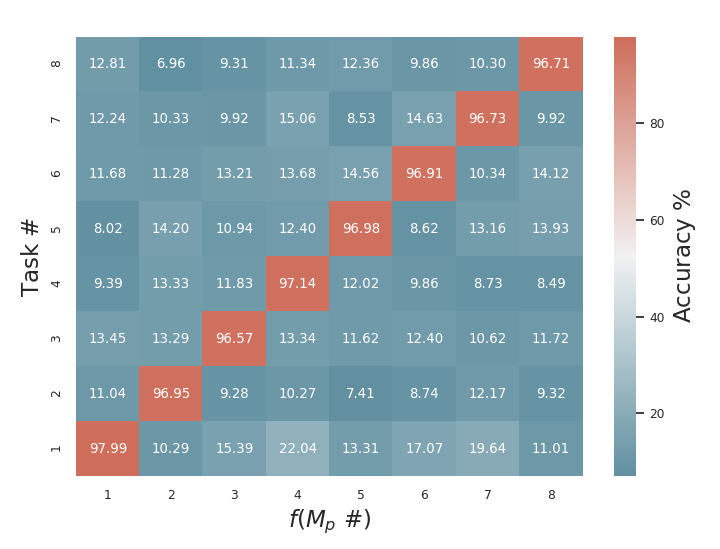}}
\vspace*{-.2in}
\caption{P-MNIST Randomized Unsupervised Task Inference}
\label{pmnist-cr2}
\end{center}
\vskip -0.35in
\end{figure}

\textbf{Setup.} After the \textit{ANN} has been trained, the test data is randomized and provided to the model for inference without its task identity (something which would happen in real world \textit{CL} scenario). The model identifies the task to which the data belongs to and then the test accuracy is calculated from the correctly identified task over the entire task set.
For \textit{RMNs}, as task $j$ is not given at inference time, thus $max_k f(x, k; W)$ is returned, as seen in Algorithm 2 (Suppl. Sec. 2).
Our experimental results show that for any ground-truth task label $j$, indeed the desired result is $f(x, j; W) \approx max_k f(x, k; W)$, which allows for unsupervised CL inference, as the pathways of different tasks don't overlap unless the tasks are the same.

\textbf{Results and Discussion.}
Unfortunately, we couldn't find any \textit{SOTA} \textit{CL} method which can be used for this experiment, or can be used with trivial modifications. It should be possible for \cite{lee_continual_2020} to possibly be able to extend the method to do unsupervised task inference. However since the method employs a mixture of expert models for every task as well as generative replay which in turn rapidly drives up computational and storage memory requirements for even small \textit{ANNs}, it cannot be considered a strict \textit{CL} setup or even a slightly relaxed version of the same. In Figure~\ref{pmnist-cr2}, we show how our algorithm is able to detect the right task - we see that the relevance-weight combination achieves the correct maximum activation in the final layer only when the correct relevance is used. We also display the percentage of correct activations for other relevance values even if they are not maximum activations. According to our knowledge, our method is the only known continual learning method under \textit{strict} \textit{CL} setup constraints capable of successfully accomplishing unsupervised task inference.

\section{Conclusion}

In this work, we study the twin problem of catastrophic forgetting and remembering in continual learning.
To resolve them, we introduce Relevance Mapping for continual learning, which applies a relevance map on the parameters of a neural network that is learned concurrently to every task. In particular, Relevance Mapping learns an optimal overlap of network parameters between sequentially learned tasks, reducing the representational overlap for dissimilar tasks, while allowing for overlap in the network parameters for related tasks.
We demonstrate that our model efficiently deals with catastrophic forgetting and remembering, and \textbf{achieves SOTA performance across a wide range of popular benchmarks} without relaxing the conditions of a strict continual learning framework.
\nocite{sri14, ando05, ebrahimi_uncertainty-guided_2020-1,oswald_continual_2019, ahn_uncertainty-based_nodate, kurle_continual_2020, knoblauch_optimal_2020, kri, Lake1332, 10027939599}
\newpage
\bibliography{references}
\bibliographystyle{icml2021}

\end{document}


\twocolumn[
\icmltitle{Supplementary}

\icmlkeywords{Machine Learning, ICML}

\vskip 0.3in
]

\section{Concepts}

\subsection{Strict Continual Learning}
In Section $1$ of the main paper, the concept of \textit{Continual Learning} is defined and it is noted that most of the current state of the art Continual Learning methods relax the constraints of a \textit{strict} continually learning framework. In Table~\ref{baselines}, we quote some of the major violations of a \textit{strict continual learning} framework with reference to the state of the art methods compared in the main paper. 

\textit{Data Replay} refers to the usage of old or future task data in any way to train the neural network. Methods like \cite{guo2020improved, fear17, nguyen2017variational, titsias2019functional, pan2021continual, Chaudhry_2018_ECCV}, etc. all employ this tactic in unique ways to learn continually. It often involves saving old task data in memory modules and been originally inspired from \cite{robins_catastrophic_1993} who was among the first researchers to show that data replay helps in alleviating catastrophic forgetting in artificial neural networks, albeit at an expense of relaxing the constraints of a \textit{strict} continual learning setup.

\textit{Multihead} usually refers to the usage of different last (usually linear) layer for each task. This has become particularly common in continual learning benchmarks with many methods~\cite{Chaudhry_2018_ECCV} dissuading the usage of single heads for a continually learning neural network. There are a few methods which also employ similar but unique methodology. For e.g.~\cite{Serr2018OvercomingCF} uses a binary hard coded final layer per task.

\textit{Pretrained} refers to using a pretrained model (usually trained on a more complex dataset like ImageNet~\cite{imagenet_cvpr09}) for training on a simpler problem or dataset (e.g. CIFAR~\cite{kri}). Having used data outside of the continual learning problem setup and since the model has now probably \textit{over-generalized} to entire sequel data/task set, using a pretrained model relaxes the constraints of a \textit{strict} continual learning.

\textit{Generative Replay} ordinarily refers to the usage of generative modelling for the dataset or task at hand and using these generated samples for some form of data replay. The usage of additional generative model (even for simple classification models and tasks), saving old data points, etc. all violate the conditions of a \textit{strict} continual setup.

\textit{Multi-Models} refers to the usage of a neural model other than the original model to help with the continual learning problem. This can reflect in using meta networks, or in strategies similar to the one presented in~\cite{yoo_snow_2020} where separate \textit{delta} models are used per task to help the original neural network learn continually.

The aforementioned concepts are not mutually exclusive and clearly do not present an exhaustive list of methods employed to relax a \textit{strict} continual learning framework, however they do provide us a reference among-st the compared \textit{SOTA} methods to ascertain which method displays the best results with least amount of constraint relaxation. 

Table~\ref{baselines} show that our method, \textit{RMNs}, do not need to use any of the aforementioned methods to violate \textit{strict continual learning} constraints and still produces the state of the art results in common Continual Learning benchmarks.
\begin{table*}[!htb]
  \caption{Common Methods used in Continual Learning which relax a \textit{strict} \textit{CL} framework }
  \label{baselines}
  \vskip 0.15in
  \begin{center}
  \begin{small}
  \begin{sc}
  \begin{threeparttable}
  \begin{tabular}{lccccc}
    \toprule
    Algorithm & Data Replay & Multihead & Pretrained & Generative Replay & Multi-Models\\
    \midrule
    VCL& \cmark & \cmark & \xmark& \xmark & \xmark \\
    HAT& \xmark & \cmark & \xmark & \xmark & \xmark\\
    RWALK& \cmark & \cmark$^*$ & \xmark& \xmark & \xmark \\
    AGS-CL &\xmark & \cmark & \xmark$^*$& \xmark & \xmark \\
    FRCL& \cmark & \cmark & \xmark& \xmark & \xmark \\
    MEGA-II& \cmark & \cmark$^*$ & \xmark& \xmark & \xmark \\
    SNOW& \xmark & \cmark & \cmark& \xmark & \cmark \\
    FROMP& \cmark & \cmark$^*$ & \xmark& \xmark & \xmark \\
    \midrule\midrule
    DLP & \xmark & \xmark & \xmark& \xmark & \xmark \\
    EWC & \xmark & \cmark$^*$ & \xmark& \xmark & \xmark \\
    SI & \xmark & \cmark$^*$ & \xmark& \xmark & \xmark \\
    MAS & \xmark & \cmark & \xmark$^*$& \xmark & \xmark \\
    \textbf{RMN (Ours)} & \xmark & \xmark & \xmark & \xmark & \xmark \\
    \bottomrule
    \end{tabular}
    \begin{tablenotes}[para]\footnotesize
    \item[*] Exceptions exist
    \end{tablenotes}
    \end{threeparttable}
    \end{sc}
    \end{small}
    \end{center}
    \vskip -0.1in
\end{table*}
\subsection{Catastrophic Remembering}
A common solution for the problem of \textit{Catastrophic Forgetting} is to \textit{overgeneralize} to the entire set of sequential data/tasks. For example, methods which employ data replay, pre-training and knowledge distillation directly employ \textit{over-generalization} for CF alleviation.

\textbf{Over-generalization}
The concept of over-generalization (in the case of back propagated artificial neural networks) refers to the learning of parameter set by the neural network tries to or has already learned a much more general function than what is required. 

A simple example to understand \textit{Catastrophic Remembering} was provided in the work by French \yrcite{french1999catastrophic} where a network has a task of reproducing an input as output. A new input is detected if output diverges by a large margin. If the network learns too well and learns the identity function, then it has \textit{overgeneralized} and hence loses the ability to detect new input. This trivial example presents one aspect of \textit{Catastrophic Remembering}. However, there is no guarantee that the loss of discrimination always leads to correct generalization - the network just becomes too familiar with the input irrespective of whether the output is correct. 


\section{Relevance Mapping Method}
\subsection{Algorithms}
\subsubsection{Supervised}
In the supervised continual learning setup, task labels are available both during training and inference (though \textit{RMNs} do not requires task labels as such). This kind of experimental setup is currently the most common form of evaluation used for Continual Learning methods. A point to note that adding regularization to induce sparsity in \textit{RMNs} is optional and is not required to obtain an optimally trained model (as shown in Section~\ref{spar}). Additionally, \textit{model weights are never pruned} in \textit{RMN} methodology. The \textit{pruning} mentioned in Algorithm~\ref{mmnts} and \ref{algo2} refers to zeroing out of relevance mappings which have not tightened towards value of $1$. The prune parameter $\mu$ may also refer to the combination of weight and $\mathbb{M}_P$.
\begin{algorithm}[H]
	\caption{\textit{RMN} Supervised Continual Learning}
	\label{mmnts}
\begin{algorithmic}[1]
    \STATE {\bfseries Input:} data $x$, ground truth $y$ for $n$ tasks, prune parameter $\mu$, corresponding task labels $i$ paired with all $x$
    \STATE {\bfseries Given:} parameters $\mathbf{W}$ \& initilaized relevance mappings $\mathbb{M_P}$
	\FOR{each task $i$}
	\STATE $f(x_i; \mathbf{W}, \mathbf{\mathbb{M}_{P_i}}) \Rightarrow \hat{y_i} = \sigma((W \odot \mathbb{M}_{P_i}) \odot x_i)$
	\STATE Compute Loss $: L(\hat{y_i},y_i)$ 
	\STATE Optional: Add Sparsity Loss : $L(\hat{y_i},y_i) + (\mathbb{M}_{P_i})_{l_0}$ 
	\STATE Backpropagate and optimize
	\STATE Prune $\mathbb{M_P}\leq \mu$ only. 
	\STATE Stabilize (fix) parameters in $f$ where $\mathbb{M_P} = 1$
	\ENDFOR
	\STATE {\bfseries Inference:} For data $x$ and ground-truth task label $i$:
	\STATE {\bfseries Output:} $f(x,i;W)$
\end{algorithmic}
\end{algorithm}

\subsubsection{Unsupervised}
For Unsupervised learning setup (which is used as a measure of Catastrophic Remembering in our work), we introduce two sub tests - new task/data detection and unsupervised/randomized task inference.

In New Task/Data Detection, task label information is unavailable during both training and inference time and the model has to detect the new task in a unsupervised manner.
\begin{algorithm}[H]
	\caption{\textit{RMN} Unsupervised Continual Learning}
	\label{algo2}
\begin{algorithmic}[1]
    \STATE {\bfseries Input:} data $x$, ground truth $y$, prune parameter $\mu$
    \STATE {\bfseries Given:} parameters $\mathbf{W}$, $\mathbb{M_P}_{est\_j}$ with $est\_j=0$, Task Switch Detection Method \textit{TSD}
	\FOR{each task $j$}
	\STATE Filter input $x$ on $f(x; \mathbf{W}, \mathbf{\mathbb{M_P}_{0...j-1}})$
	\STATE $f(x; \mathbf{W}, \mathbf{\mathbb{M_P}_{est\_j}}) \Rightarrow \hat{y}$
	\STATE Compute Loss : $L(\hat{y},y)$ 
	\IF{$TSD(x)$ is True}
	\STATE $est\_j++$
	\STATE Add $\mathbb{M_P}_{est\_j}$ to learn-able parameter list
	\STATE $f(x; \mathbf{W}, \mathbf{\mathbb{M_P}_{est\_j}}) \Rightarrow \hat{y}$
	\STATE Re-Compute Loss : $L(\hat{y},y)$
	\ENDIF
	\STATE Backpropagate and optimize
	\STATE Sample $x_g$ from standard Gaussian distribution with same shape as $x$
	\STATE $f(x_g; \mathbf{W}, \mathbf{\mathbb{M_P}_{est\_j}}) \Rightarrow \hat{y}$
	\STATE Compute Loss : $||\hat{y}-0||_2^2$ 
	\STATE Backpropagate and optimize
	\STATE Prune $\mathbb{M_P}\leq \mu$ only. 
	\STATE Stabilize (fix) parameters in $f$ where $\mathbb{M_P} \approx 1$
	\ENDFOR
	\STATE {\bfseries Inference:} For data $x$:
	\STATE {\bfseries Output:} $max_k f(x,k;W)$
\end{algorithmic}
\end{algorithm}
This usually involves task boundary detection during training which often involves statistical tests like Welch's T-test or KL-Divergence on the model's loss~\cite{titsias2019functional, pan2021continual}. However, this setup often assumes clean sequential data over the entire training set as well as in a mini-batch. New Task/Data detection also involves learning in an unsupervised way when the incoming mini-batch is noisy i.e. a mix of old and new task data. The Fuzzy Unsupervised Learning experiment done on Sequential-MNIST follows the procedure laid out in~\cite{lee_continual_2020}.

\textit{The Randomized Unsupervised Task Inference} experiment also capitalizes on the aforementioned weakness of the former test. Under this, a continually trained model has to identify the task ID during inference under circumstances where the inference input is task randomized. However, since \textit{RMNs} form unique subnetworks in the original neural network, they are capable of trivially identifying the task ID as well as filtering out the noisy training data for new tasks.
\subsection{$\beta$ parameter}
As presented in the main text, we used a sigmoidal pseudo-round function during training which is completely rounded during inference: \begin{equation}
\mathcal{L}_R(x_k; \beta) = \frac{1}{1 + \exp(-(\beta (x_k - 0.5)))}
\end{equation}

We also noted that $\lim_{\beta \longrightarrow \infty} (\mathcal{L}_R(x, \beta))$ for $x \in [0,1]$ is equivalent to the rounding function. Here, $\beta$ is a learnable, layer-wise parameter (i.e., in our implementation, there is one specific $\beta$ for every layer of a given network) that controls the ``tightness'' of $\mathcal{L}_R$. We experimented with different values of $\beta$ and noted that for an arbitrary high value of $\beta$ ($\geq 80$), there's not any visible difference in results and that the $\beta$ value doesn't require any tuning. If we tried and learn the $\beta$ parameter instead of fixing its value, we noted that the $\beta$ value tightened over time.

\subsection{Sparsity Analysis}\label{spar}
\begin{figure}
\vskip 0.2in
\begin{center}
\centerline{\includegraphics[width=\columnwidth]{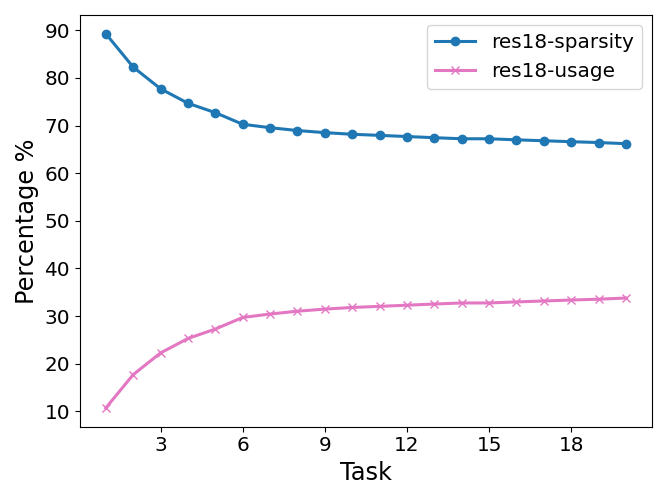}}
\vspace{-\baselineskip}
\caption{Model Sparsity for AMN-Resnet-18 trained on CIFAR-100 (20 tasks)}
\label{scifar}
\end{center}
\end{figure}

Sparsity($\mathcal{S}$) in a \textit{RMN} $f(\mathcal{W}, \mathbb{M}_P)$ is calculated according to the following formula\begin{equation}
    \mathcal{S} = \frac{\prod_{i=1}^n\cap\mathbb{M}_P = 0}{num(\mathcal{W})}
\end{equation}
Figure~\ref{scifar} shows the model sparsity and usage for Resnet-18 model trained on Cifar-100~\cite{kri} dataset over $20$ tasks. For this experiment, \textit{there are no loss functions, regularization or any method employed to constrain the model parameters or $\mathbb{M}_P$ for sparsity.} If we do however constrain for sparsity using $L_1$ or $L_0$ regularization, we can observe much higher sparsity levels. We can observe that the model capacity usage evens out over time which can be explained due to subsequent tasks finding overlap amongst old task parameters.

\subsection{Model Computational Complexity w.r.t number of Tasks}
\textit{RMNs} require only the learned weights of the continually learning network, though this is achieved through creating distinct sub-network mappings in each network. This does increase the number of parameters, but ultimately reduces the effective model size because all additional parameters can be converted to binary tensors. Thus, the memory complexity can be written as ${O(tk)}$ where $t$ is an integer and equal to the number of tasks and $k$ is a constant. Thus, for a finite and \textit{constrained} value of $t$, the memory complexity of \textit{RMNs} is $O(1)$ i.e. constant. The value of $k$ depends on the amount of overlap in our model as well as the method used to save binary parameters. For e.g. For the model in Table~\ref{mlp}, the theoretical worst case scenario (a model with no overlap amongst the relevance mappings) results in $12$kb of memory. Practically, as noticed in the \textit{RMN} sparsity discussion, the model has not been observed to fully utilize its weights over the period of sequential tasks and unused parameters can be effectively removed post training. Additionally, we do not implement bias parameters in our \textit{RMN}. Thus, effectively, the final model memory footprint is actually \textit{negative} as compared with even the baseline model for all the experiments.

\subsection{The Lottery Ticket Hypothesis and Relevance Mapping}
A question arises as to whether the slight constraints introduced in the weight space by our algorithm worsen the performance of the sequential tasks. The Lottery ticket Hypothesis introduced in a seminal work~\cite{frankle2019lottery} states that - \textit{A randomly-initialized, dense neural network contains a subnetwork that is initialized such that—when trained in isolation—it can match the test accuracy of the original network after training for at most the same number of iterations.} Additionally, our method doesn't remove the previous tasks parameters and subsequent methods can choose to use their predecessors parameters optimally. Therefore, no performance drop is expected in our method and results from our experiments prove the same.


\section{Experimental Details}\label{exp}
The experimental implementation for most comparative methods mentioned in Table~\ref{baselines} have been taken from official implementations of \cite{Serr2018OvercomingCF}, \cite{titsias2019functional}, \cite{jung2020continual} and \cite{pan2021continual}.
\subsection{Architectural Details}

In this section, we provide detailed descriptions of the architectures used for our experiments. We denote 2D convolutional layers as Conv2D, linear layers as Linear, Rectified Linear Unit as ReLU, and Batch Normalization as BN. For \textit{RMN} versions of each layer with included Relevance Mapping $\mathbb{M_P}$, we add an ``M-'' prefix, e.g. M-Conv is the \textit{RMN} version of Conv.

For fair comparison purposes, for all architectures, we attempt to keep architectural representational capacity and module sequences as similar as possible to referenced methods.

\begin{table}[h]
  \caption{Details of the Multi-layer Perceptron network used for the Permuted-MNIST and Split-MNIST tasks. This architecture is of equivalent representational capacity to the network used in \cite{titsias2019functional}.}
  \label{mlp}
  \vskip 0.15in
  \begin{center}
  \begin{small}
  \begin{sc}
  
  \begin{tabular}{c}
    \toprule
    Input: $x \in \mathbb{R}^{784}$ \\
    \midrule
    M-Linear $(784) \longrightarrow 100$\\
    \midrule
    M-BN (100)\\
    \midrule
    ReLU \\
    \midrule
    M-Linear $(100) \longrightarrow 100$\\
    \midrule
    M-BN $(100)$\\
    \midrule
    ReLU \\
    \midrule
    M-Linear $(100) \longrightarrow 100$\\
    \midrule
    M-BN $(100)$\\
    \midrule
    ReLU \\
    \midrule
    m-Linear $(100) \longrightarrow 10$\\
    \midrule
    SoftMax\\
    \bottomrule
  \end{tabular}
  \end{sc}
  \end{small}
  \end{center}
  \vskip -0.1in
\end{table}

\begin{table}[h]
  \caption{Details of the convolutional network used for the Sequential Omniglot task. This architecture is of similar or equivalent representational capacity to the network used in \cite{titsias2019functional}, cited as ``Baseline'' in their results sections.}
  \label{mlp2}
  \vskip 0.15in
  \begin{center}
  \begin{small}
  \begin{sc}
  
  \begin{tabular}{c}
    \toprule
    Input: $x \in \mathbb{R}^{1\times105\times105}$\\
    Resize $\longrightarrow$ $x \in \mathbb{R}^{1\times28\times28}$\\
    \midrule
    M-Conv2D $1 ch \longrightarrow 250 ch$\\
    \midrule
    M-BN (250)\\
    \midrule
    ReLU \\
    \midrule
    MaxPooling (2 x 2), $stride = 2$\\
    \midrule
    M-Conv2D $250 ch \longrightarrow 250 ch$\\
    \midrule
    M-BN (250)\\
    \midrule
    ReLU \\
    \midrule
    MaxPooling (2 x 2), $stride = 2$\\
    \midrule
    M-Conv2D $250 ch \longrightarrow 250 ch$\\
    \midrule
    M-BN (250)\\
    \midrule
    ReLU \\
    \midrule
    MaxPooling (2 x 2), $stride = 2$\\
    \midrule
    M-Conv2D $250 ch \longrightarrow 250 ch$\\
    \midrule
    M-BN (250)\\
    \midrule
    ReLU \\
    \midrule
    MaxPooling (2 x 2), $stride = 2$\\
    \midrule
    m-Linear $\longrightarrow 60$\\
    \midrule
    SoftMax\\
    \bottomrule
    \end{tabular}
    \end{sc}
    \end{small}
    \end{center}
    \vskip -0.1in
\end{table}
Tablea~\ref{mlp} and \ref{mlp2} provide the main architecture details of the continual learning experiments dealing with Sequential MNIST, Permuted MNIST and Sequential Omniglot benchmarks.

For the sequential Cifar-100 (10 tasks)~\cite{kri} benchmark, we follow the experimental and architectural details from \cite{jung2020continual}. For the \textit{RMN}, however, we do not use bias parameters and dropout layers. Table~\ref{cnet} shows the architecture details for the \textit{RMN} used in the experiment.

For the \textit{RES-CIFAR} experiment which uses a Resnet-18~\cite{he2015deep}\footnote{HAT~\cite{Serr2018OvercomingCF} authors mentioned that there is no official implementation for Residual Networks} trained over Sequential Cifar-100 (20 tasks), we use the original Resnet-18 architecture for all experiments with modifications as required by a specific method\footnote{AGS-CL~\cite{jung2020continual} authors did not reply to our query concerning official Residual network implementation.}. For the \textit{RMN}-Resnet-18, we do not make use of bias parameters and dropout layers.

\begin{table}[h]
  \caption{Details of the convolutional network used for the Sequential Cifar 100 (10 tasks) task. This architecture is of similar or equivalent representational capacity to the network used in \cite{jung2020continual}}
  \label{cnet}
  \vskip 0.15in
  \begin{center}
  \begin{small}
  \begin{sc}
  
  \begin{tabular}{c}
    \toprule
    Input: $x \in \mathbb{R}^{3\times32\times32}$\\
    \midrule
    M-Conv2D $3 ch \longrightarrow 32 ch$\\
    \midrule
    ReLU \\
    \midrule
    M-Conv2D $32 ch \longrightarrow 32 ch$\\
    \midrule
    ReLU \\
    \midrule
    MaxPooling (2 x 2), $stride = 2$\\
    \midrule
    M-Conv2D $32 ch \longrightarrow 64 ch$\\
    \midrule
    ReLU \\
    \midrule
    M-Conv2D $64 ch \longrightarrow 64 ch$\\
    \midrule
    ReLU \\
    \midrule
    MaxPooling (2 x 2), $stride = 2$\\
    \midrule
    M-Conv2D $64 ch \longrightarrow 128 ch$\\
    \midrule
    ReLU \\
    \midrule
    M-Conv2D $128 ch \longrightarrow 128 ch$\\
    \midrule
    ReLU \\
    \midrule
    MaxPooling (2 x 2), $stride = 2$\\
    \midrule
    M-Linear $\longrightarrow 256$\\
    \midrule
    M-Linear $\longrightarrow 10$\\
    \bottomrule
    \end{tabular}
    \end{sc}
    \end{small}
    \end{center}
    \vskip -0.1in
\end{table}

\subsection{Hyperparameter Details}

In this section, we provide detailed descriptions of training, optimization, and hyperparameter details.

In \cite{titsias2019functional} and \cite{jung2020continual}, they select experiment with a range of hyperparameters, choose the values that return the highest validation accuracy and then retrain on the union of the train and validation set. When applicable, we select hyperparameter values similar or equivalent to those arrived at in \cite{titsias2019functional} for MNIST and Omnliglot experiments and \cite{jung2020continual} for Cifar-100 experiments.
For all continual learning tasks, we make use of the Adam optimizer and have separate learning rates for weights and $\mathbb{M_P}$ parameters. Subsets of weights are frozen via gradient masking as tasks increase, where $\prod_t^T\mathbb{M_{P_t}}=1 $ is the mask applied to the weights at task $T+1$. 

For the Permuted-MNIST and Split-MNIST tasks, we use a 90-10 train-test split, 0.002 learning rate for all parameters, and batch size of 128. For all tasks, the network is trained for 250 epochs.  

For the Sequential Omniglot task, we use an 80-20 train-test split, 0.0002 learning rate for all parameters, except $\mathbb{M_P}$ parameters, a learning rate of 0.0001 for $\mathbb{M_P}$ parameters, a batch size of 16. For the first task, the network is trained for 150 epochs, for subsequent tasks the network trained for 200 epochs.

In S-CIFAR-100 and RES-CIFAR, we train all comparative methods with mini-batch size of $256$ for $100$ epochs using Adam optimizer\cite{kingma2014adam} with initial learning rate $0.001$ and decaying it by a factor of $3$ if there is no improvement in the validation loss for $5$ consecutive epochs, similarly as in \cite{jung2020continual, Serr2018OvercomingCF}. 

For our method \textit{(RMNs)}, we keep the same mini batch size, training epochs and optimizer as mentioned in \cite{jung2020continual}. For Split Cifar-100 (10 tasks) and RES-CIFAR (Split Cifar100-20 tasks with Resnet-18), the model weight parameters initial learning rate is .001 and for $\mathbb{M}_P$ the learning rate is .01. The prune parameter value is .05 and .01 respectively which is used to prune the relevance mappings. The pruning is done whenever the model's task loss converges which varies from epoch $20-80$ for different tasks.

\section{Similar Methods}
\subsection{Differences with Relevance Mapping Method}
\cite{Serr2018OvercomingCF} proposes hard attention (\textit{HAT}), a task based attention mechanism which can be considered the most similar to our \textit{RMN}. 

It differs from \textit{RMN} due to following reasons- 
\begin{enumerate}
    \item They utilize task embeddings and a positive scaling parameter - and a gated product of these two is used to produce a non-binary mask - unlike our \textit{RMNs} which don't use either a task embedding or a scaling parameter and is necessarily binary.
    \item Unlike \textit{RMNs}, the attention on the last layer in \textit{HAT} is manually hard-coded for every task.
    \item A recursive cumulative attention mechanism is employed to deal with multiple non binary mask values over tasks in \textit{HAT}. \textit{RMNs} however have no need for such a mechanism.
    \item \textit{HAT} cannot be used in a unsupervised \textit{CL} setup or to deal with \textit{CR} and has not been implemented with more complex network architectures like Residual Networks.
\end{enumerate}

\cite{jung2020continual} uses \textit{proximal gradient descent algorithm} to progressively freeze nodes in an \textit{ANN}. 
\begin{enumerate}
    \item Unlike \textit{RMNs}, this method employs selective regularization to signify node importance (which is calculated by lasso regularization). 
    \item This method progressively uses up the parameter set of the \textit{ANN} and it is unclear whether it can be used for an arbitrary large number of sequential tasks. 
    \item This method employs two group sparsity-based penalties in order to regularize important nodes, however \textit{AMN} do not require usage such kind of sparse based penalty.
    \item This method is also unable to deal with unsupervised learning scenario or \textit{CR}. (iv) This method uses a different classification layer for each task - relaxing the core constraints of the problem altogether.
\end{enumerate}

\cite{ferrari_memory_2018} 
\begin{enumerate}
    \item Calculates the parameter importance by calculating sensitivity of the squared $l2$ norm of the function output to their changes and then uses regularization (similar to \cite{kirkpatrick2017overcoming} to enforce in sequential learning, unlike \textit{RMNs}. 
    \item The method enforces fixed synaptic importance between tasks irrespective to their similarity and unlike our work, doesn't seem to be capable of working under Unsupervised Learning scenarios.
\end{enumerate}
    
\cite{yoo_snow_2020} propose \textit{SNOW} and 
\begin{enumerate}
    \item Uses a unique channel pooling scheme to evaluate the channel relevance for each specific task which differs from \textit{RMN's} individual node relevance mapping strategy. 
    \item Importantly, this work, unlike \textit{RMNs}, employs a pre-trained model which is frozen source model which already \textit{overgeneralizes} to the \textit{CL} problem at hand and thus makes this method inapplicable for dealing with \textit{CR}. 
    \item It also doesn't seem to be capable of handling unsupervised learning/testing scenarios.
\end{enumerate}

\newpage
\bibliography{supplementary_references}
\bibliographystyle{icml2021}


\appendix